\documentclass[journal]{IEEEtran}

\usepackage{times}
\usepackage{epsfig}
\usepackage{graphicx}
\usepackage{subfigure}
\usepackage{amsmath}
\usepackage{amssymb}
\usepackage{amsfonts}
\usepackage{bm}
\usepackage{color}

\begin{document}

\title{A Deep Information Sharing Network for Multi-contrast Compressed Sensing MRI Reconstruction}

\author{Liyan Sun$^{\ast}$,~ Zhiwen Fan$^{\ast}$,~ Yue Huang,~ Xinghao Ding,~ John Paisley$^{\dagger}$\\
\thanks{This work was supported in part by the National Natural Science Foundation of China under Grants 61571382, 81671766, 61571005, 81671674, U1605252, 61671309 in part by the Guangdong Natural Science Foundation under Grant 2015A030313007, in part by the Fundamental Research Funds for the Central Universities under Grant 20720160075, 20720180059, in part by the National Natural Science Foundation of Fujian Province, China under Grant 2017J01126. (Corresponding author: Xinghao Ding)}
\thanks{L. Sun, Z. Fan, Y. Huang and X. Ding was with the School of Information Science and Engineering, Xiamen University, Xiamen, Fujian, 361005, China, corresponding to: dxh@xmu.edu.cn.}% <-this % stops a space
\thanks{$^{\dagger}$ J. Paisley was with Department of Electrical Engineering, Columbia University, New York, NY, USA}% <-this % stops a space
\thanks{$^{\ast}$ The first co-authors contributed equally.}% <-this % stops a space
}

\markboth{Journal of \LaTeX\ Class Files,~Vol.~14, No.~8, August~2015}%
{Shell \MakeLowercase{\textit{et al.}}: Bare Demo of IEEEtran.cls for IEEE Journals}

\maketitle

\begin{abstract}
  In multi-contrast magnetic resonance imaging (MRI), compressed sensing theory can accelerate imaging by sampling fewer measurements within each contrast. The conventional optimization-based models suffer several limitations: strict assumption of shared sparse support, time-consuming optimization and ``shallow'' models with difficulties in encoding the rich patterns hiding in massive MRI data. In this paper, we propose the first deep learning model for multi-contrast MRI reconstruction. We achieve information sharing through feature sharing units, which significantly reduces the number of parameters. The feature sharing unit is combined with a data fidelity unit to comprise an inference block. These inference blocks are cascaded with dense connections, which allows for information transmission across different depths of the network efficiently. Our extensive experiments on various multi-contrast MRI datasets show that proposed model outperforms both state-of-the-art single-contrast and multi-contrast MRI methods in accuracy and efficiency. We show the improved reconstruction quality can bring great benefits for the later medical image analysis stage. Furthermore, the robustness of the proposed model to the non-registration environment shows its potential in real MRI applications.
\end{abstract}


\IEEEpeerreviewmaketitle

\section{Introduction}

\IEEEPARstart{M}{agnetic} resonance imaging has been widely used to generate anatomically precise images of in-vivo tissue. A major limitation of MRI is the relatively slow data acquisition speed. Compressed sensing (CS)  has therefore been used to accelerate MRI by reducing the number of the k-space (i.e., Fourier) measurements directly acquired by the machine \cite{1}. CS theory shows how accurate or even perfect reconstruction can be achieved via appropriate optimizations to fill in the missing Fourier coefficients of k-space \cite{2}. Recently, the compressed sensing MRI has been approved by FDA to two main MRI vendors: GE and Siemens \cite{38}. Hence more MRI scans are expected to be produced using compressed sensing methods in clinic, where maintaining the high reconstruction quality with rapid imaging speed is important to improve the performance of later analysis stage and patients' comfort. The compressed sensing for magnetic resonance imaging (CS-MRI) is also an active research topic in medical imaging, and now one of the classic inverse imaging problems in the field of computer vision.

Similar to other tasks in image restoration and reconstruction, research on CS-MRI is driven by proposing an effective optimization model for MRI reconstruction. For example, MRI is modeled by sparsity constraints in a fixed transform bases, e.g., SparseMRI \cite{5}, TVCMRI \cite{6}, RecPF \cite{7} and FCSA \cite{8,9}. Limited by the representation ability of the models with non-adaptive transform basis, some work devoted to utilizing the geometric information within image patches such as PBDW \cite{10}, PANO \cite{11}, FDLCP \cite{12} and GBRWT \cite{13}. Ravishankar and Bresler \cite{14} and Huang et al. \cite{15} also introduced dictionary learning into CS-MRI.

\begin{figure}
\begin{center}
   \subfigure[PD MRI] { \label {figure1a} \includegraphics[width=0.15\textwidth]{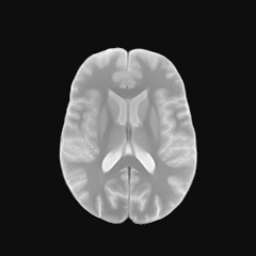}}
   \subfigure[T1 MRI] { \label {figure1b}\includegraphics[width=0.15\textwidth]{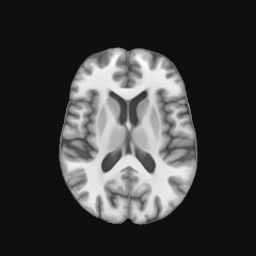}}
   \subfigure[T2 MRI] {\label {figure1c} \includegraphics[width=0.15\textwidth]{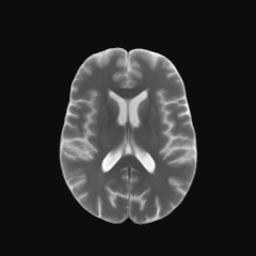}}
   \subfigure[TV for PD] { \label {figure1d} \includegraphics[width=0.15\textwidth]{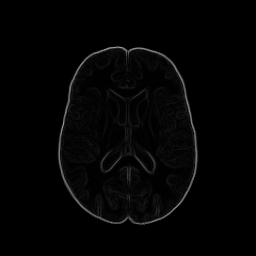}}
   \subfigure[TV for T1] { \label {figure1e} \includegraphics[width=0.15\textwidth]{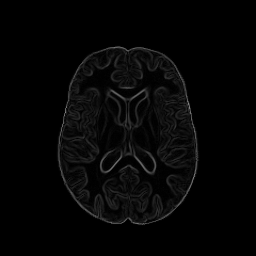}}
   \subfigure[TV for T2] { \label {figure1f} \includegraphics[width=0.15\textwidth]{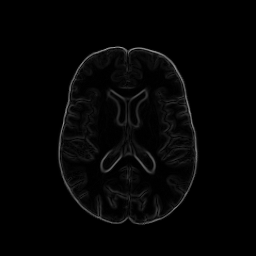}}
   \caption{Multi-contrast MRI images share similar structures.}
\label {figure1}
\end{center}
\end{figure}

As \cite{13,14} show, models with adaptive transform bases achieve higher reconstruction quality, but at the expense of heavy computational burden. Furthermore, conventional optimization-based CS-MRI methods are implemented \textit{in situ}, meaning they do not rely on information from MRI training data. The first issue is a clear drawback, while the second may have positive aspects, but the power of deep learning has shown a clear advantage in exploiting big data resources with a deep model.

Thus, deep neural networks have recently been introduced to CS-MRI. For example, Wang et al. \cite{16} use a vanilla CNN model to learn the mapping from zero-filled MRI to fully-sampled MRI via a massive MRI training set. (Note the term ``zero-filled MRI'' means the missing Fourier coefficients are replaced by zeros, followed by an inverse 2D FFT.) Sun et al. \cite{17} proposed ADMM-NET as a modification of the alternating direction method of multipliers (ADMM) algorithm where the parameters are inferred via back-propagation. Lee et al. \cite{18} proposed a modified U-Net to learn the mapping in the residual domain. Notably, Schlemper et al. \cite{19,20} proposed a \emph{deep cascade convolutional neural network} (DC-CNN) to unroll the standard paradigm of CS-MRI into the deep learning architecture. The DC-CNN represents the state-of-the-art performance in \textit{single-contrast} CS-MRI in both imaging quality and speed.

The work mentioned above is based on single-contrast CS-MRI reconstruction. Usually, an MRI scan can obtain images of the same anatomical section under different contrasts, such as T1, T2, and proton-density (PD) weighted MRI generated, by applying different MRI protocols \cite{21}. Multi-contrast MRI contains similar but not the same image structures. By comparing multiple contrast MRI in the same region, radiologists can detect subtle abnormalities such as a developing tumor. This is illustrated in Figure \ref{figure1a}, \ref{figure1b} and \ref{figure1c}, where the PD, T1 and T2 MRI in the SRI24 \cite{35} datasets exhibit similar structures. In the second row we show the root of sum of square of the horizontal and vertical gradients of the multi-contrast MR images. Rather than reconstruct each multi-contrast MRI independently, joint reconstruction can provide higher quality images by exploiting such structural similarity.

In this paper, we propose the first deep models for multi-contrast CS-MRI reconstruction. We start with two basic networks called \emph{deep independent reconstruction network} (DIRN) and \emph{deep feature sharing network} (DFSN). DIRN uses separate parallel networks to reconstruct each contrast of the MRI with each network a state-of-the-art DC-CNN architecture \cite{19}. DFSN takes the further step of applying a feature sharing strategy that significantly reduces the number of network parameters. Our final deep model, which extends the state-of-the-art results of DFSN, uses a dense connection strategy to transfer information across layers in the network. We call this end-to-end model a \emph{deep information sharing network} (DISN) for multi-contrast CS-MRI inversion. DISN comprises cascaded and densely connected inference blocks consisting of feature sharing units and data fidelity units. In the feature sharing units, all multi-contrast MRI share the same feature maps. We use dense connections to help information sharing at different depths.
%The additional parameters brought by the dense connection increase in linear way, and we prove the DISN can achieve better performance than DFSN with less network parameters. Besides, we retrain the DISN model under non-registration environment, and the experiment results on the test data show the DISN is robust to non-registration and has potential in real MRI applications.

 Our contributions can be summarized as follows:

 \begin{itemize}
   \item In the proposed basic DFSN model, the feature sharing unit fully exploits the similarity among the multi-contrast MRI. The comparative experiments show the DFSN model outperforms DIRN model with multiple amounts of parameters of the independent parallel networks.
   \item In the proposed DISN model, the dense connection operation is proposed to propagate the information from lower blocks to deeper blocks directly. The number of parameters only increase linearly rather than quadratically in the regular DenseNet \cite{33}. Even with much fewer network parameters, the dense connection strategy still shows advantages.
   \item The experiments on various multi-contrast MRI datasets show the proposed DISN model achieves the state-of-the-art performance compared with both single-contrast and multi-contrast MRI methods in imaging quality and speed. We also show the improved reconstruction quality can significantly benefit the later analysis stage.
   \item The DISN model is robust to the non-registration situations because of large model capacity, which is the usual case in real MRI application.
 \end{itemize}

The rest of this paper is organized as follows: Section \uppercase\expandafter{\romannumeral2} provides the related work in the field of multi-contrast MRI reconstruction. Section \uppercase\expandafter{\romannumeral3} elaborates the basic DIRN and DFSN model and the proposed DISN models. Section \uppercase\expandafter{\romannumeral4} compares the different deep models and reports the experimental results on various multi-contrast MRI datasets including SRI24 \cite{35}, MRBrainS13 \cite{40} and NeoBrainS12 \cite{39}. Section \uppercase\expandafter{\romannumeral5} discusses the network size, testing running time, non-registration environment. Finally in Section \uppercase\expandafter{\romannumeral6} we draw the conclusions.

\begin{figure*}
\begin{center}
   \subfigure[The network architecture of DIRN network] { \label {figure2a} \includegraphics[width=1\textwidth]{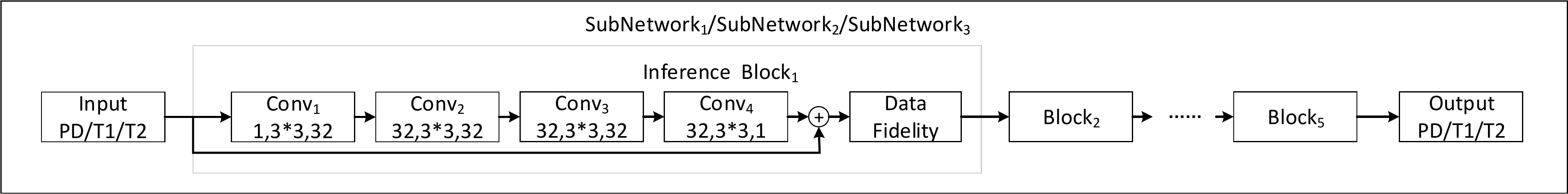}}
   \subfigure[The network architecture of DFSN network] { \label {figure2b} \includegraphics[width=1\textwidth]{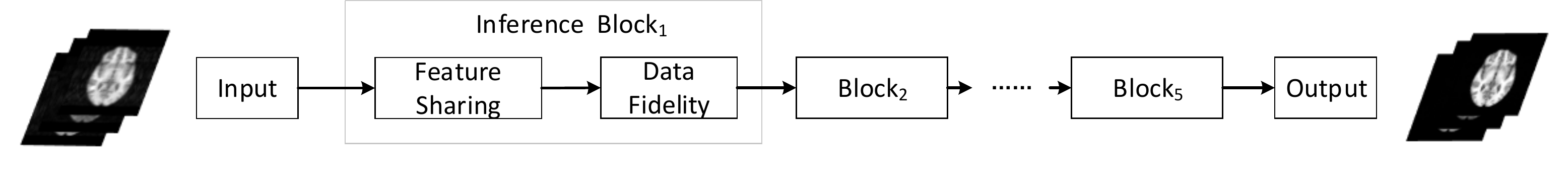}}
   \subfigure[The network architecture of DISN network] { \label {figure2c} \includegraphics[width=1\textwidth]{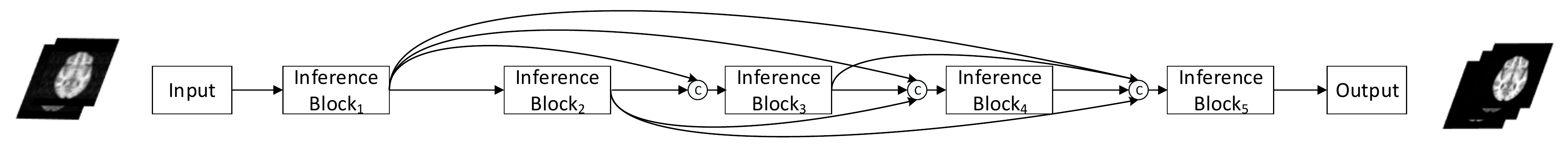}}
   \caption{The network architecture of the proposed DIRN, DFSN and DISN models for multi-contrast CS-MRI inversion.}
\label {figure2}
\end{center}
\end{figure*}

\section{Related Works: Compressed sensing for Multi-contrast MRI Reconstruction}
\label{RelatedWorks}

Previous work has exploited the structural correlations in multi-contrast MRI using non-deep approaches. Suppose we aim at reconstructing $L$ multi-contrast MRI images, for example $L = 3$ when PD, T1 and T2 MRI are used. One can formulate this problem as
\begin{equation}\label{eq1}
X = \mathop {\arg \min }\limits_X \sum\limits_{i = 1}^L {\frac{{{\lambda _i}}}{2}\left\| {{F_{{u_i}}}{x_i} - {y_i}} \right\|_2^2}  + \rho \left( X \right),
\end{equation}
where $x_i \in {\mathbb C^{N \times 1}}$ denotes the $i^{th} \left( {1 \le i \le L} \right)$ contrast of the complex-valued MR image to be reconstructed and $X$ indicates the set of all $x_i$. ${F_{{u_i}}}\in {\mathbb C^{M \times N}}$ denotes the $i^{th}$ under-sampled Fourier matrix and $y_i\in {\mathbb C^{M \times 1}}$ ($M<N$) denotes the $i^{th}$ k-space measurements. Note that in the field of multi-contrast MRI, it is common to under-sample all the multi-contrast MRI data using different under-sampling masks with the same under-sampling ratio. The first term is called the data fidelity and ensures consistency between the reconstructed image and measurements. ${\rho}\left( X \right)$ encodes a regularization for the MRI contrast images.

Two notable approaches to multi-contrast CS-MRI with which we compare are Bayesian Compressed Sensing by Bilgic et al. \cite{23} and FCSA-MT by Huang et al. \cite{26}.

\subsection{Bayesian Compressed Sensing}

Bilgic et al. base their approach on a modification to Bayesian compressed sensing (BCS) \cite{24} that exploits structural similarity across contrasts.
To exploit the structure similarity, the author cast the problem in gradient domain, the vertical and horizontal gradients of the multi-contrast MRI are set zero-mean Gaussian distributions as prior. The prior distributions for each contrast share the same precision estimated in maximum likelihood fashion by the MRI with all the contrasts. According the conjugacy, the posteriors of the gradients are also Gaussian distributions. The inferred gradients are used to produce the reconstruction for each contrast via least square.
Combining the horizontal and vertical gradients with the k-space data fidelity, a least squared problem can be solved to yield the reconstruction.

We note the all the multi-contrast MRI images contribute to the estimation of the precision of the gradients and the precision is shared.
Instead of imposing strict sparse support assumption among these multi-contrast MRI, the BCS model controls the similarity by expressing uncertainty, while also allowing for the idiosyncrasy in each contrast. However, the BCS method suffers several limitations: (1) The sparsity is imposed on gradient domain, which is an improved variant of total variation regularization in essence. (2) Each coefficients in gradient domain is imposed on a unimodal Gaussian distribution, which is difficult to capture the diversity patterns in MRI images. (3) More importantly, the running time of the BCS algorithm is long, eg., about 26 hours for processing a set of multi-contrast MRI data.
The running time of the initial algorithm is impracticable in real scenarios, although the authors later accelerated the model at the expense of performance \cite{25}.

\subsection{Group Sparsity}

Huang et al. extended the FCSA \cite{8,9} algorithm designed originally for single-contrast MRI to multi-contrast MRI called FCSA-MT \cite{26,27}.
The FCSA-MT model is based on two key observations: 1) Across the multi-contrast MRIs, the variance of the gradients should show similarity in the same spatial positions. 2) The wavelet coefficients across the multi-contrast MRIs also should have similar non-zero supports in the same anatomical sections. In the FCSA-MT model, the least sqaured data fidelity fitting with joint total variation (JTV) regularization and group wavelet sparsity regularization is proposed as the loss function,
\begin{equation}
\begin{aligned} \label{eq2}
 x = \mathop {\arg \min }\limits_x \frac{1}{2}\sum\nolimits_{i = 1}^L {\left\| {{F_{{u_i}}}X\left( {:,i} \right) - {y_i}} \right\|_2^2}  \\
 + \alpha {\left\| X \right\|_{JTV}} + \beta {\left\| {\Phi X} \right\|_{2,1}},
\end{aligned}
\end{equation}
where the vectorized multi-contrast MRI images are arranged in column to form the data matrix $X$, and the joint total variation is defined as ${\left\| X \right\|_{JTV}} = \sum\nolimits_{i = 1}^N {\sqrt {\sum\nolimits_{s = 1}^L {\left( {{{\left( {{\nabla _1}{X_{is}}} \right)}^2} + {{\left( {{\nabla _2}{X_{is}}} \right)}^2}} \right)} } }$ and the wavelet group sparsity regularized in the form of $\ell_{2,1}$ norm ${\left\| {\Phi X} \right\|_{2,1}} = \sum\nolimits_{i = 1}^N {\sqrt {\sum\nolimits_{s = 1}^L {{{\left( {\Phi {X_{is}}} \right)}^2}} } }$, where $\nabla_1$ and $\nabla_2$ stands for horizontal and vertical difference operator and $\Phi$ stands for the orthogonal wavelet transform matrix. The FCSA-MT model achieves balance between the model performance and computational efficiency.
In this approach, structural correlations are modeled as group sparsity, which can clearly outperform its single-contrast FCSA counterpart on this problem.
However the model is limited by the fixed transform domain, eg., finite difference and wavelet and the model is designed in situ, no external data is used to provide further information. Besides, the group sparsity assumption that the multi-contrast MRI have similar sparse supports in $\ell_2$ way is too strict, especially under non-registration environment.
Huang et al. later accelerated FCSA-MT using fast conditioning \cite{28}.

Compared with the BCS and FCSA-MT models, the proposed DISN model can encode the complex patterns within the multi-contrast MRI datasets. After training stage, the forward pass is highly efficient because no iteration for optimization is required. Another major advantage over previous methods is the powerful non-linear mapping ability, overcoming the strict sparse support assumption, which is valuable in real applications in MRI.

%Some other work in the multi-contrast MRI method fullsample at least one MRI image while keep other under-sampled. The full-sampled MRI is set reference image to provide information for the reconstruction for other undersampled MRI. Ehrhardt and Betcke \cite{26} proposed the structure-guided total variation (SGTV) regularization for multi-contrast MRI reconstruction. Qu et al. \cite{27} utilized the guided graph representation provided by reference or full-sampled MRI to help the reconstruction. Weizman et al. \cite{28} proposed the reference-based MRI for multi-contrast MRI. While these work demands high imaging quality of the reference image and are limited in real MRI application.

\section{A deep information sharing network (DISN)}
\label{SecDISN}

We propose a deep model that takes a set of sub-sampled Fourier k-space measurements at multiple contrasts, $y_1,\dots,y_L$, and outputs the corresponding image reconstructions at each contrast $x_1,\dots,x_L$. The model learns how to exploit structural similarities across these contrasts to produce an output that is significantly better than can be obtained via $L$ independent inversion algorithms. Because this represents the first deep learning approach to the problem, we experiment with three different deep structures, but the best-performing structure is a ``deep information sharing network'' (DISN). This consists of cascaded blocks with dense connections. Within each block, we adopt a feature sharing unit combined with a data fidelity unit. Below, we elaborate on the feature sharing unit and data fidelity unit, as well as how they are combined to form the inference block. We then discuss how the blocks are connected densely in an efficient way.

\subsection{Feature sharing unit}

\begin{figure}
\begin{center}
    { \includegraphics[width=1\columnwidth]{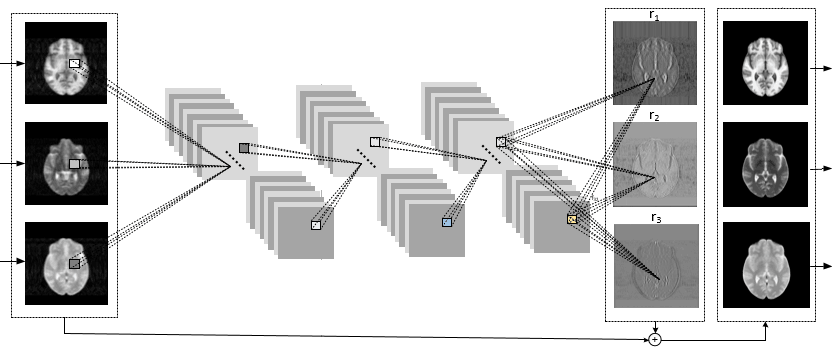}}
   \caption{The feature sharing unit.}
\label {figure3}
\end{center}
\end{figure}

An intuitive approach to multi-contrast CS-MRI inversion is to simply reconstruct them separately with a deep model, eg., as shown in Figure \ref{figure2a}. We call this a \emph{deep independent reconstruction network} (DIRN). The DIRN model we show is made up of several parallel subnetworks. Here we plot $3$ subnetworks for PD, T1 and T2 contrasts. The architecture of each subnetwork is the state-of-art DC-CNN architecture \cite{19}. If each subnetwork consists of $N$ inference blocks, we name it DIRN-$N$B (All the subnetworks share the same number of blocks). Here we adopt $5$ blocks for example, eg. , DIRN-$5$B. Each building block consists of $4$ convolutional layers with global shortcut and a data fidelity unit (we will discuss this later). In each block, the first convolutional layer is used to map the MRI images to multiple feature maps and the last convolutional layer integrates the feature maps into a single reconstruction in residual domain. The Leaky ReLU is used as activation function
except for the last
convolutional layer, where the identity mapping is used. There are no interactions among these subnetworks. In such a deep network setting, massive MRI data is used to learn the complex patterns of each multi-contrast MRI separately.

Although DIRN may provide powerful modeling ability for each contrast of the MRI, the number of network parameters is tripled because if there are three subnetworks. As we showed in Figure \ref{figure1}, the structural similarity should be exploited in deep neural network architectures, both to achieve better reconstruction and also with the aim of reducing parameters. Hence we also consider a \emph{deep feature sharing network} (DFSN) as shown in Figure \ref{figure2b}. Similar to DIRN, the DFSN consists of $5$ cascaded inference blocks, eg. , DFSN-$5$B, while each block is made up of a feature sharing unit and a data fidelity unit. The multi-contrast like T1, T2 and PD zero-filled MR magnitude images are fed to the DFSN in a stack. The DFSN network can therefore reconstruct multi-contrast MRI data simultaneously. We show the feature sharing unit in Figure \ref{figure3}. In traditional multi-contrast MRI methods, the structural
similarity is modeled in the finite difference domain; instead we adopt \textit{residual learning} in the feature sharing unit. Similarly, each feature sharing unit contains $4$ convolutional layers with the same number of filters as the single subnetwork of DIRN in each layer and all activation functions are Leaky ReLU, except fore the identity mapping in the last convolutional layer.

\begin{figure}
\begin{center}
    { \includegraphics[width=1\columnwidth]{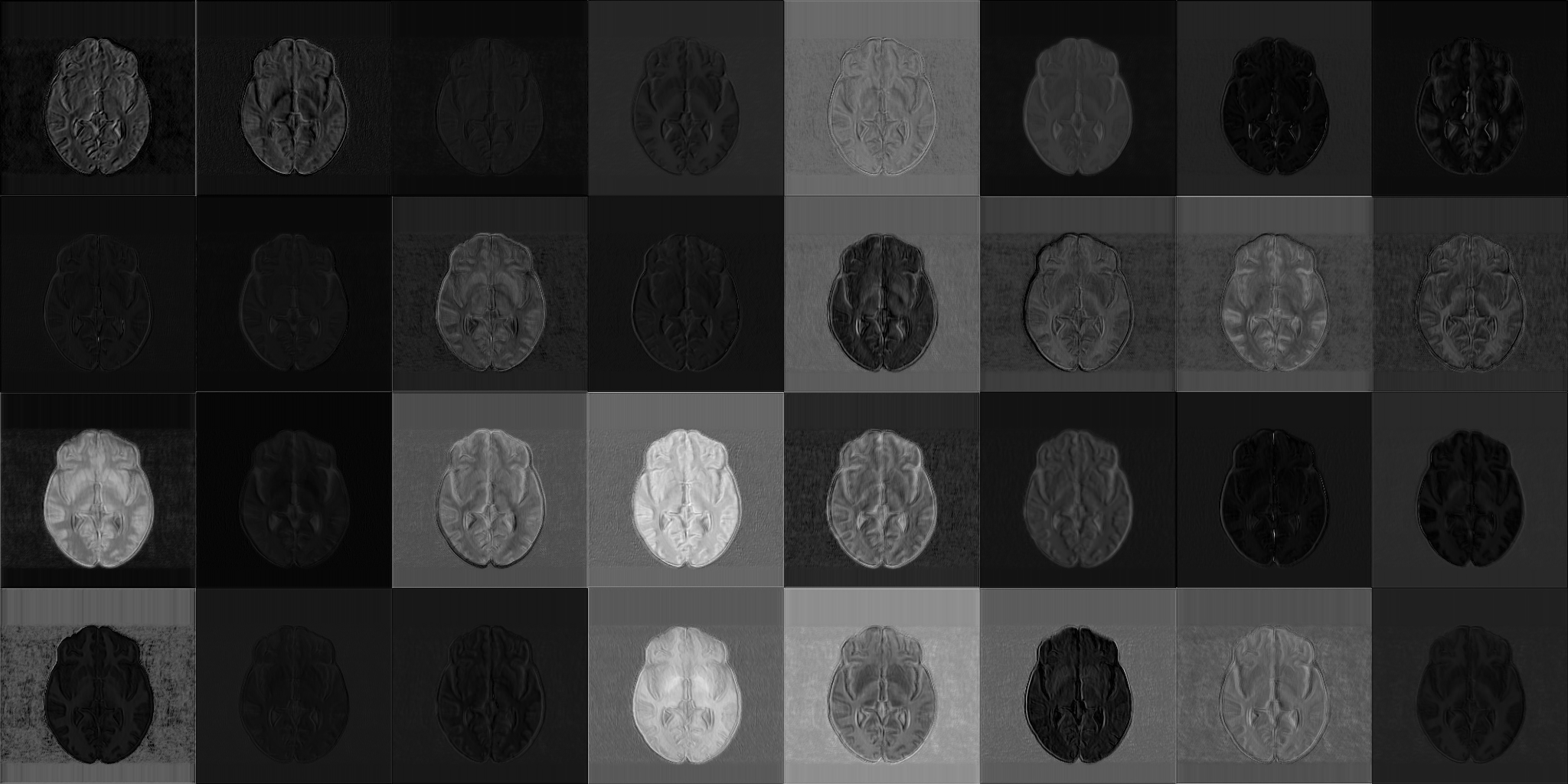}}
    \caption{The feature maps from the last convolutional layer in the feature sharing unit within the last block of the DFSN-5B model.}
\label {figure4}
\end{center}
\end{figure}

\subsubsection{Discussion}
The proposed feature sharing strategy has a similar motivation to traditional sparse representation methods. In each feature sharing unit, we denote the residuals for the $j^{th} \left( {1 \le j \le L} \right)$ contrast MRI as $r_j$. As mentioned previously, for the last convolutional layer in the unit, the activation function is set to the identity function, thus the following equation holds: ${r_j} = \sum\nolimits_i {{f_i}{w_{ij}}}$, where the $f_i$ denotes the $i^{th}$ feature map for the last convolutional layer in the unit, and $w_{ij}$ denotes the corresponding kernel in Toeplitz matrix form.  In the classic dictionary learning formulation, the signal can be approximated as $s = D\alpha$, or equivalently $s = \sum\nolimits_i {{d_i}{\alpha _i}}$. $d_i$ is the $i^{th}$ column of the dictionary $D$ and $\alpha_i$ is the $i^{th}$ entry of the sparse coefficients $\alpha$.

In previous work such as the ScSR model for image super-resolution \cite{29,30}, the patches of high and low
resolution images share the same sparse coefficients $\alpha_j$ yet different dictionaries $D_h$ and $D_l$. In such setting the correlation between the low and high resolution image patches may be overlooked. While in the DFSN model, the part function similar to high-resolution and low-resolution dictionaries, in the form of $f_i$, are inferred simultaneously with the representation coefficients ${w_{ij}}$ via a large dataset. In Figure \ref{figure4}, we show the feature maps $f_i \left( {1 \le i \le 32} \right)$ for the last convolutional layer in the feature sharing unit of the $5^{th}$ block with the DFSN-$5$ model. We observe that they contain enough diversity to represent the structures in PD, T1 and T2 MRI, which validates the feature sharing strategy.
% In the field of image processing of color images, usually the RGB channels are input simultaneously into the network instead of processed separately, which explicitly utilize their correlations \cite{33}.

Besides, in contrast to FCSA-MT \cite{27} or BCS \cite{23}, where the ``dictionary" is fixed or learned \textit{in situ}, the transform basis in the DFSN model is inferred via massive MRI data pairs, which fully utilizes the rich patterns hiding in the data.

\subsection{Data fidelity unit}

We also use the data fidelity unit \cite{19} within each block to reduce bias by enforcing more accurate values on the sampled positions in k-space. Following Equation \ref{eq1}, we solve the following objective function in the data fidelity unit for each contrast,
\begin{equation}\label{eq3}
{x_i} = \mathop {\arg \min }\limits_{{x_i}} \frac{\lambda_i }{2}\left\| {{F_{{u_i}}}{x_i} - {y_i}} \right\|_2^2 + \frac{1}{2} \left\| {{x_i} - {x_{i{n_i}}}} \right\|_2^2,
\end{equation}
where $x_{in_i}$ is the input to the data fidelity unit and $\lambda_i$ is the regularization parameter. To enforce consistency between the reconstruction and the measurements $y_i$, we set $\lambda_i$ a large value, e.g., $10^6$, which only penalizes deviation from these measured locations. The second term can be viewed as the prior guess, where the input image $x_{in_i}$ is the output by the feature sharing unit. We observe that these fidelity units are calculated independently for each contrast, but each input $x_{in_i}$ is constructed by sharing information across contrasts in the deep model.

We can simplify by working in the Fourier domain, after which the solution is (using element-wise division below)
\begin{equation}\label{eq4}
{x_i} = {F^H}\frac{{\lambda FF_{{u_i}}^H{y_i} + F{x_{i{n_i}}}}}{{\lambda FF_{{u_i}}^H{F_{{u_i}}}{F^H} + I}},
\end{equation}
where the term $FF_{u_i}^H{y_i}$ is the Fourier transform of the zero-filled reconstruction, the term $FF_{u_i}^H{F_{u_i}}{F^H}$ is a diagonal matrix with ones at the sampled locations and zeros otherwise.
Calling the feed forward function for this unit $g\left( {{x_{in_i}};{y_i};\lambda } \right)$, the relevant gradient for model training is
\begin{equation}\label{eq5}
\frac{{\partial g}}{{\partial {x_{in_i}}^T}} = \frac{I}{{\lambda FF_{u_i}^H{F_{u_i}}{F^H} + I}}.
\end{equation}

\begin{figure}
\begin{center}
   \subfigure[inter-recon from B1] { \label {figure5a} \includegraphics[width=0.15\textwidth]{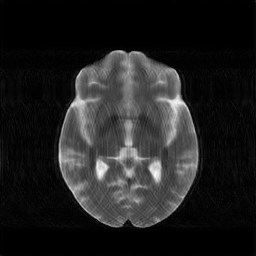}}
   \subfigure[inter-recon from B2] { \label {figure5b} \includegraphics[width=0.15\textwidth]{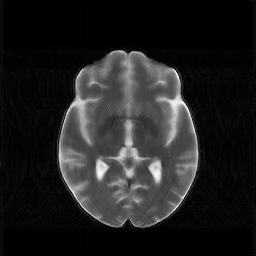}}
   \subfigure[inter-recon from B3] { \label {figure5c} \includegraphics[width=0.15\textwidth]{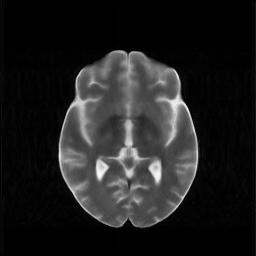}}
   \subfigure[inter-recon from B4] { \label {figure5d} \includegraphics[width=0.15\textwidth]{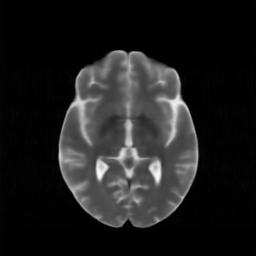}}
   \subfigure[inter-recon from B5] { \label {figure5e} \includegraphics[width=0.15\textwidth]{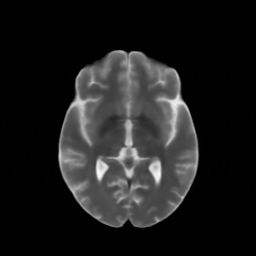}}
   \subfigure[reference]           { \label {figure5f} \includegraphics[width=0.15\textwidth]{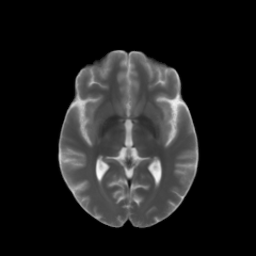}}\\
   \subfigure[\scriptsize map B1]  { \label {figure5g} \includegraphics[width=.173\columnwidth]{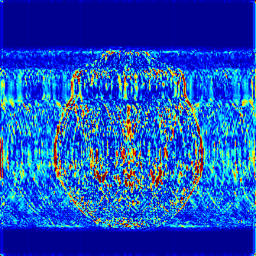}}
   \subfigure[\scriptsize map B2]  { \label {figure5h} \includegraphics[width=.173\columnwidth]{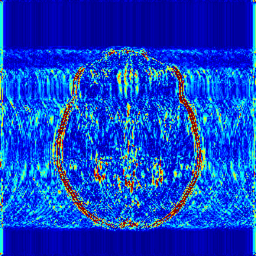}}
   \subfigure[\scriptsize map B3]  { \label {figure5i} \includegraphics[width=.173\columnwidth]{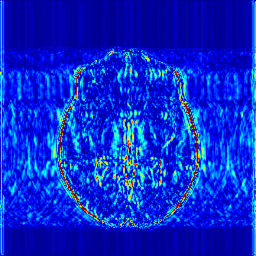}}
   \subfigure[\scriptsize map B4]  { \label {figure5j} \includegraphics[width=.173\columnwidth]{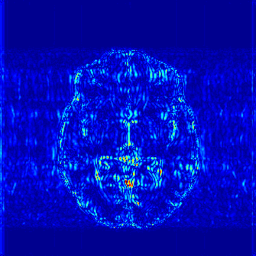}}
   \subfigure[\scriptsize map B5]  { \label {figure5k} \includegraphics[width=.173\columnwidth]{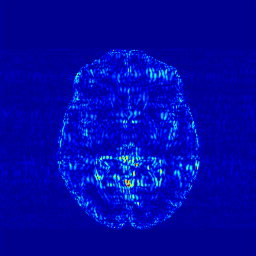}}\\
   \subfigure[\scriptsize B1-B2]   { \label {figure5l} \includegraphics[width=.173\columnwidth]{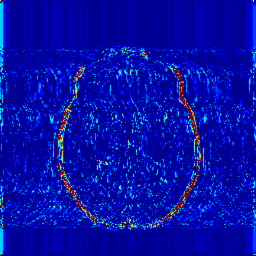}}
   \subfigure[\scriptsize B1-B3]   { \label {figure5m} \includegraphics[width=.173\columnwidth]{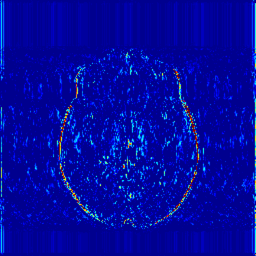}}
   \subfigure[\scriptsize B1-B4]   { \label {figure5n} \includegraphics[width=.173\columnwidth]{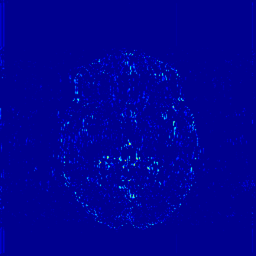}}
   \subfigure[\scriptsize B1-B5]   { \label {figure5o} \includegraphics[width=.173\columnwidth]{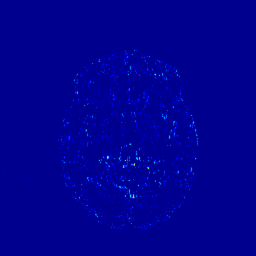}}
   \subfigure[\scriptsize B2-B3]   { \label {figure5p} \includegraphics[width=.173\columnwidth]{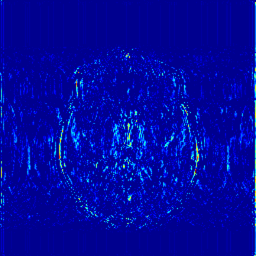}}\\
   \subfigure[\scriptsize B2-B4]   { \label {figure5q} \includegraphics[width=.173\columnwidth]{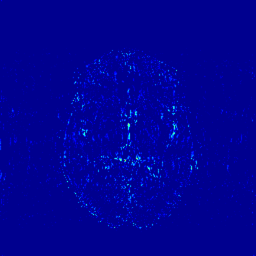}}
   \subfigure[\scriptsize B2-B5]   { \label {figure5r} \includegraphics[width=.173\columnwidth]{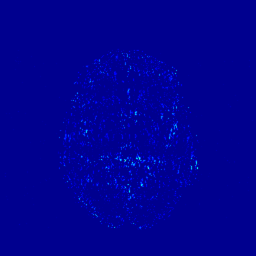}}
   \subfigure[\scriptsize B3-B4]   { \label {figure5s} \includegraphics[width=.173\columnwidth]{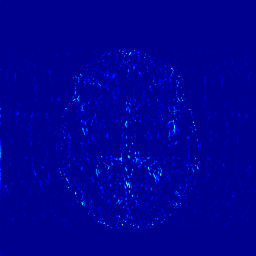}}
   \subfigure[\scriptsize B3-B5]   { \label {figure5t} \includegraphics[width=.173\columnwidth]{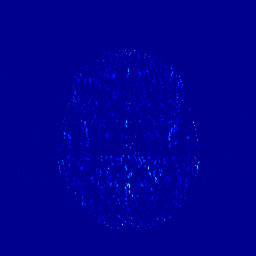}}
   \subfigure[\scriptsize B4-B5]   { \label {figure5u} \includegraphics[width=.173\columnwidth]{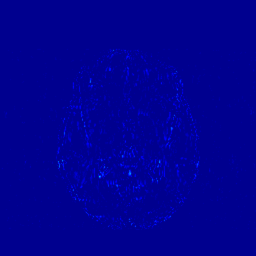}}
   \caption{The intermediate reconstructions and their error maps from each block, as well as the error differential maps.}
\label {figure5}
\end{center}
\end{figure}

\begin{figure*}
\centering
   { \label {figure5a} \includegraphics[height=1.6in]{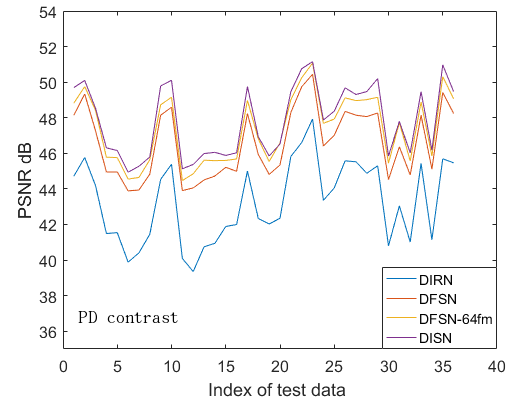}}
   { \label {figure5c} \includegraphics[height=1.6in]{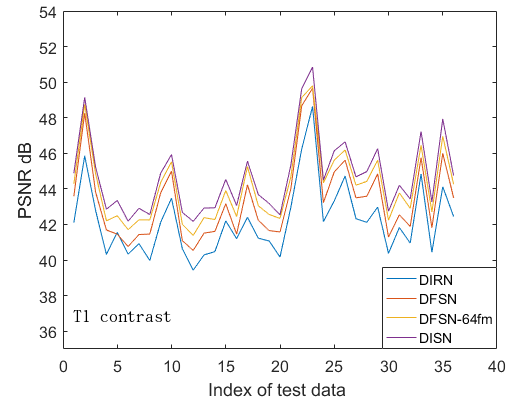}}
   { \label {figure5e} \includegraphics[height=1.6in]{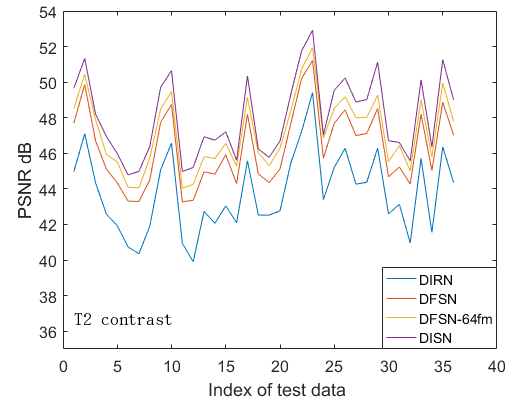}}
   { \label {figure5b} \includegraphics[height=1.6in]{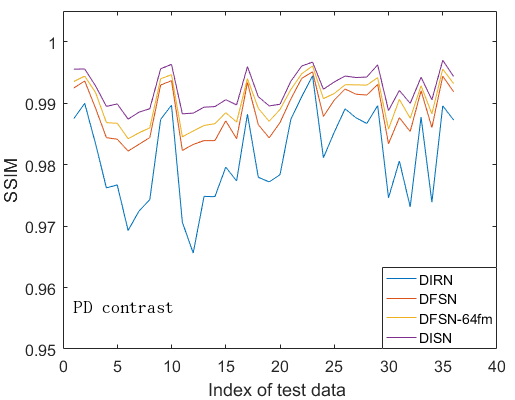}}
   { \label {figure5d} \includegraphics[height=1.6in]{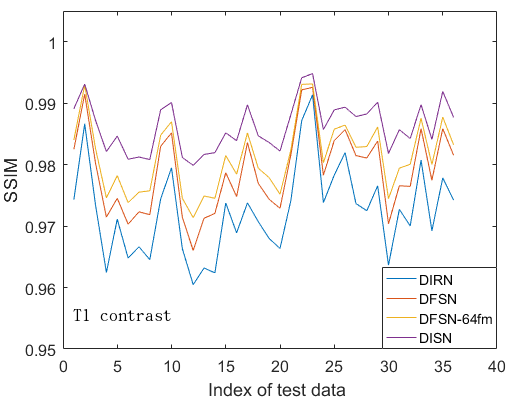}}
   { \label {figure5f} \includegraphics[height=1.6in]{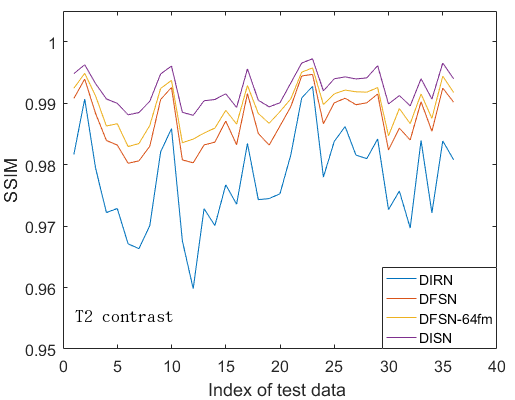}}
   \caption{The performance of DIRN-5B, DFSN-5B, DFSN-5B with 64 feature maps and DISN-5B on all $36$ test MRI in SRI24 datasets (x-axis).}
\label {figure6}
\end{figure*}

\subsection{Dense connections}

We proposed DFSN to share information across contrasts of the MRI. We visualize the intermediate reconstructions of T2 brain MRI by the inference blocks of DFSN-$5$B model in Figure \ref{figure5}. The test data is under-sampled with 1D $20\%$ Cartesian mask given in Figure \ref{figure7}. We observe that the outputs of the inference blocks keep on improving as the network goes deeper. However, we conjecture that intermediate reconstructions from lower blocks contain valuable information lost to deeper levels. We plot the pixel-wise absolute reconstruction error map of the T2 intermediate reconstruction of the DFSN-$5$B model from the first to the fifth block in the third row of Figure \ref{figure5}. In the fourth and fifth rows, we show positive error map differences of lower blocks with higher blocks, meaning we only focus on the positions where the lower block achieves high reconstruction accuracy. As the blocks go deeper, we observe the intermediate reconstructions from lower blocks show
less advantages.

Inspired by this observation, we densely connect the inference blocks in DFSN and propose a \emph{deep information sharing network} (DISN). In Figure \ref{figure2c}, we show the network architecture of the DISN-5B model. The ``information sharing'' concept is expressed in two ways: (1) The information between the multi-contrast MRI is shared via the feature sharing unit. (2) The information in the lower inference blocks and deeper inference blocks is shared by dense connections using concatenations. Each block in DISN receives the output from all previous blocks as its input.

As with DenseNet \cite{33}, where the feature maps are densely fed from lower to deeper layers by concatenation, the dimension of the channels in deeper layers may explode quadratically, limiting the depth of the model. Inspired by DenseNet and the similar MemNet \cite{34}, the DISN  is different in that only the intermediate reconstructed MRI images are concatenated rather than the large number of feature maps.
% To given an example, for DISN-5B the first and second block receive only 3 intermediate reconstruction inputs. While the third block receives 6 ones, the fourth receives 9 ones and fitth receives 12, and \etc.
As a consequence, the dimensionality only increases linearly in the channel according to the number of contrast.
% Besides, the MemNet \cite{36} is inspired by the biological phenomenon while the proposed DISN is motivated by the information sharing strategy.

\section{Experiments}

\subsection{Datasets}

We conduct experiments on three multi-contrast MRI datasets including the SRI24 atlas \cite{35}, MRBrainS13 benchmark \cite{40} and NeoBrainS12 benchmark \cite{39}.

\subsubsection{SRI24 Atlas}

The multi-contrast brain MRI atlas data was obtained on a 3.0T GE scanner with 8-channel head coil with three different contrast setting:
\begin{itemize}
  \item For T1-weighted MRI data: 3D axial IR-prep SPoiled Gradient Recalled (SPGR), TR = $6.5$ms, TE = $1.54$ms, number of slices = $124$, slice thickness = $1.25$mm.
  \item For T2-weighted (late echo) and PD-weighted (early echo) MRI data: 2D axial dual-echo fast spin echo (FSE), TR = $10$s, TE = $14/98$ms, number of slices = $62$, slice thickness = $2.5$mm.
\end{itemize}
The field-of-view covers a region of $240 \times 240$mm with resolution $256 \times 256$ pixels. The SRI24 dataset contains $407$ T1w-T2w-PD MRI training pairs. We randomly select 36 multi-contrast MRI data pairs as test data, while the others are used for training.

\subsubsection{MRBrainS}

Twenty fully annotated multi-contrast (T1-weighted, T1-weighted inversion recovery and T2-FLAIR) 3T MRI brain scans with the size $240 \times 240$ are provided in the Grand Challenge on MR Brain Image Segmentation (MRBrainS) workshop at the MICCAI2013. The voxel size of T1, T1-IR and T2-FLAIR MRI is $0.958mm \times 0.958mm \times 3.0mm$, $0.958mm \times 0.958mm \times 3.0mm$ and $0.958mm \times 0.958mm \times 3.0mm$ respectively. These scans have been acquired at the UMC Utrecht (the Netherlands) of patients with diabetes and matched controls (with increased cardiovascular risk) with varying degrees of atrophy and white matter lesions (age $>$ 50). These abnormalities have different appearances in different contrasts. The 20 scans contains 320 pairs of multi-contrast MRI data in total, we randomly select $80\%$ of the slices for training and the others for testing.

\subsubsection{NeoBrainS}

Different from the MRBrainS data which is acquired on the adult, the grand challenge in MICCAI2012 called Neonatal Brain Segmentation (NeoBrainS) provides the multi-contrast (T1-weighted and T2-weighted) MRI scans of the neonatal brains.
All the 7 scans containing 175 multi-contrast MRI data pairs of the size $512\times512$ are acquired using Philips 3T system at the University Medical Center Utrecht, The Netherlands. The detailed imaging parameters can be found in \cite{39}. We also randomly select $80\%$ of the slices as training datasets and $20\%$ as testing datasets.

\subsection{Training and parameter details}

\begin{figure}
\begin{center}
   \includegraphics[width=0.9\columnwidth]{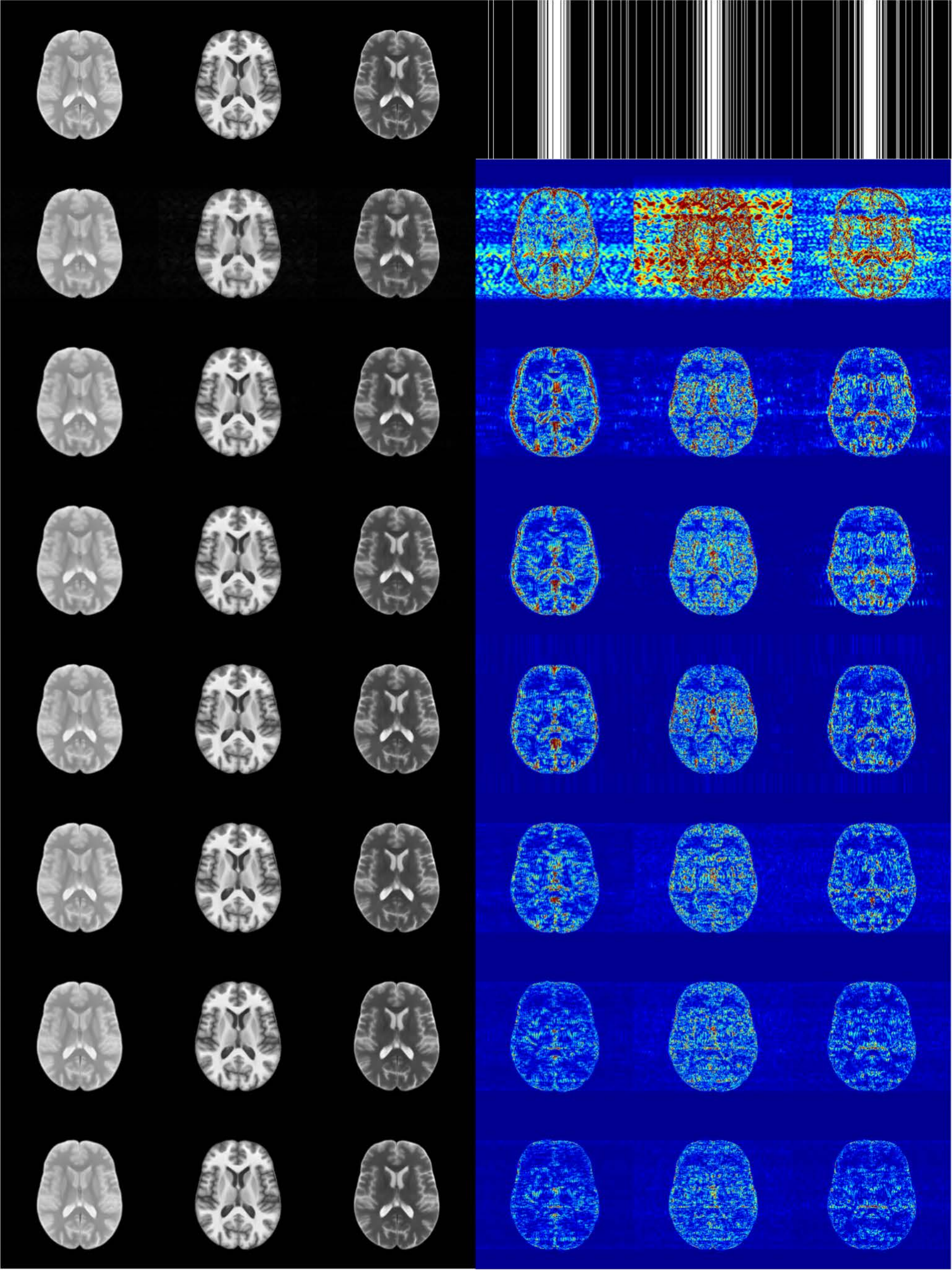}
   \caption{The reconstruction results: the first row is the fully-sampled MRI with different 1D Cartesian $20\%$ masks. From the second to eighth rows we show BCS, PANO, GBRWT, FCSA-MT, DIRN-5B, DFSN-5B and DISN-5B on a testing multi-contrast MRI data in SRI24 datasets. From left to right, each row contains the PD, T1 and T2 reconstructions and error images. The display of the error maps range from 0 to 0.1.}
\label {figure7}
\end{center}
\end{figure}

The loss function for DIRN, DFSN and DISN is
\begin{equation}\label{eq6}
L\big( {x_i^{zf},x_i^{fs}} \big) = \frac{1}{3}\sum\nolimits_{i = 1}^L {\big\| {x_i^{fs} - {f_i}\big( {x_i^{zf};{\theta _i}} \big)} \big\|_2^2}
\end{equation}
where $x_i^{zf}$ and $x_i^{fs}$ are zero-filled and fully-sampled magnitude MRI, respectively, for the $i^{th}$ contrast. The $\theta_i$ represents the network parameters for each subnetwork for DIRN, while in DFSN and DISN, they are incorporated in the single feature sharing unit.

During training, the deep models are implemented on TensorFlow for the Python environment on a NVIDIA Geforce GTX 1080Ti with 11GB GPU memory and Intel Xeon CPU E5-2683 at 2.00GHz. We use data augmentation on the entire training set. For each block within the feature sharing unit, we adopt $4$ convolutional layers followed by Leaky ReLU activation function with $0.2$ negative slope except for the last convolutional layer, where identity mapping is applied. For each convolutional layer, we obtain $32$ shared feature maps except for the first and last convolutional layer, where $3$ feature maps are used for the contrast residuals. These settings are applied to both DFSN and DISN. For the DIRN model, the first and last convolutional layer in the inference block has only one feature map since the $3$ contrasts are reconstructed using $3$ different deep networks. The kernel size is set to $3\times3$ and padding is used to keep the size of feature maps unchanged. We apply Xavier initialization for all models. We train for $40000$ iterations using ADAM. We select the initial learning rate to be 0.0005, the first-order momentum to be 0.9 and the second momentum to be 0.999. Each mini-batch contains $4$ MRI data pairs.

\subsection{Deep model comparison}

We compare DIRN, DFSN and DISN to check the utility of the feature sharing and dense connection strategies on the SRI24 atlas datasets. In Figure \ref{figure6}, we show the network performance on DIRN-$5$B, DFSN-$5$B and DISN-$5$B using all the test data, whose orders have been shuffled. We use the under-sampling pattern in Figure \ref{figure7}. Compared with DIRN, DFSN shows the advantage of feature sharing, while DISN has further improvement by exploiting information across depths by the dense connections. The number of parameters of DFSN-$5$B is $101K$, much fewer than DIRN-$5$B having $286K$, while DISN-$5$B has slightly more parameters ($106K$) because of extra kernels used in the dense connections. We also ran the experiment with $64$ feature maps for DFSN-$5$B, resulting in $387K$ parameters and found that DISN-$5$B still achieves higher reconstruction quality with $32$ feature maps. In Figure \ref{figure6} we show all results for $32$ maps and one result for $64$ maps.

\subsection{The model comparisons on the SRI24 datasets}

\begin{figure}
\begin{center}
   \subfigure[PSNR] { \label {figure8a} \includegraphics[width=0.48\textwidth]{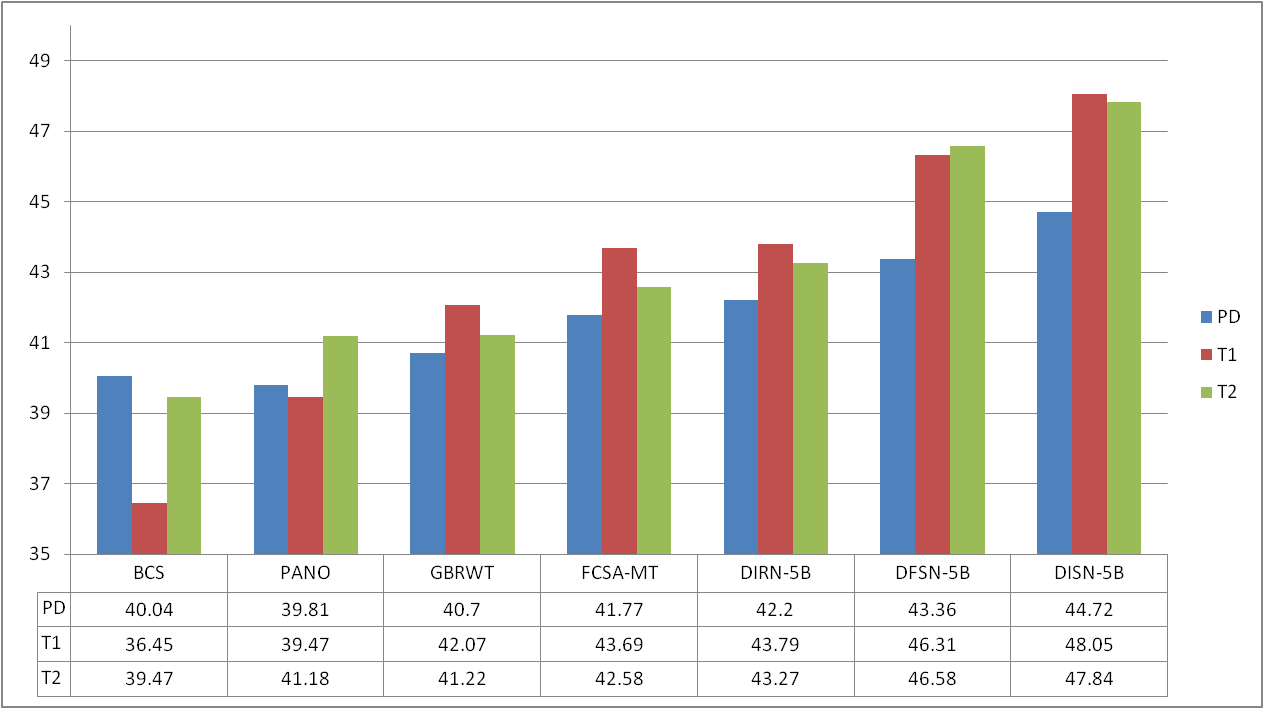}}
   \subfigure[SSIM] { \label {figure8b} \includegraphics[width=0.48\textwidth]{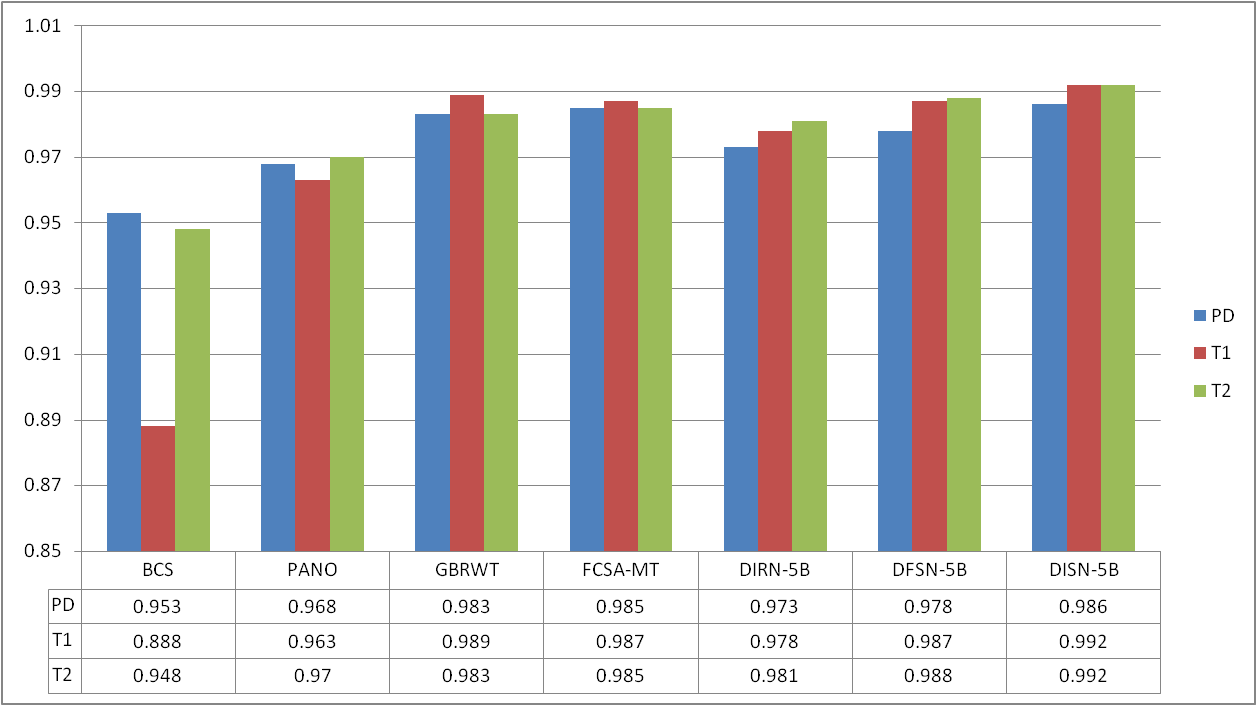}}
   \caption{The PSNR and SSIM of single- and multi-contrast CS-MRI inversion algorithms averaged over the 36 test images under the different 1D Cartesian $20\%$ masks. Deep models clearly outperform in PSNR due to their L2 minimization objective. Structural similarity (SSIM) is more competitive.\vspace{-20pt}}
\label {figure8}
\end{center}
\end{figure}

\begin{figure}
   \includegraphics[width=.99\columnwidth]{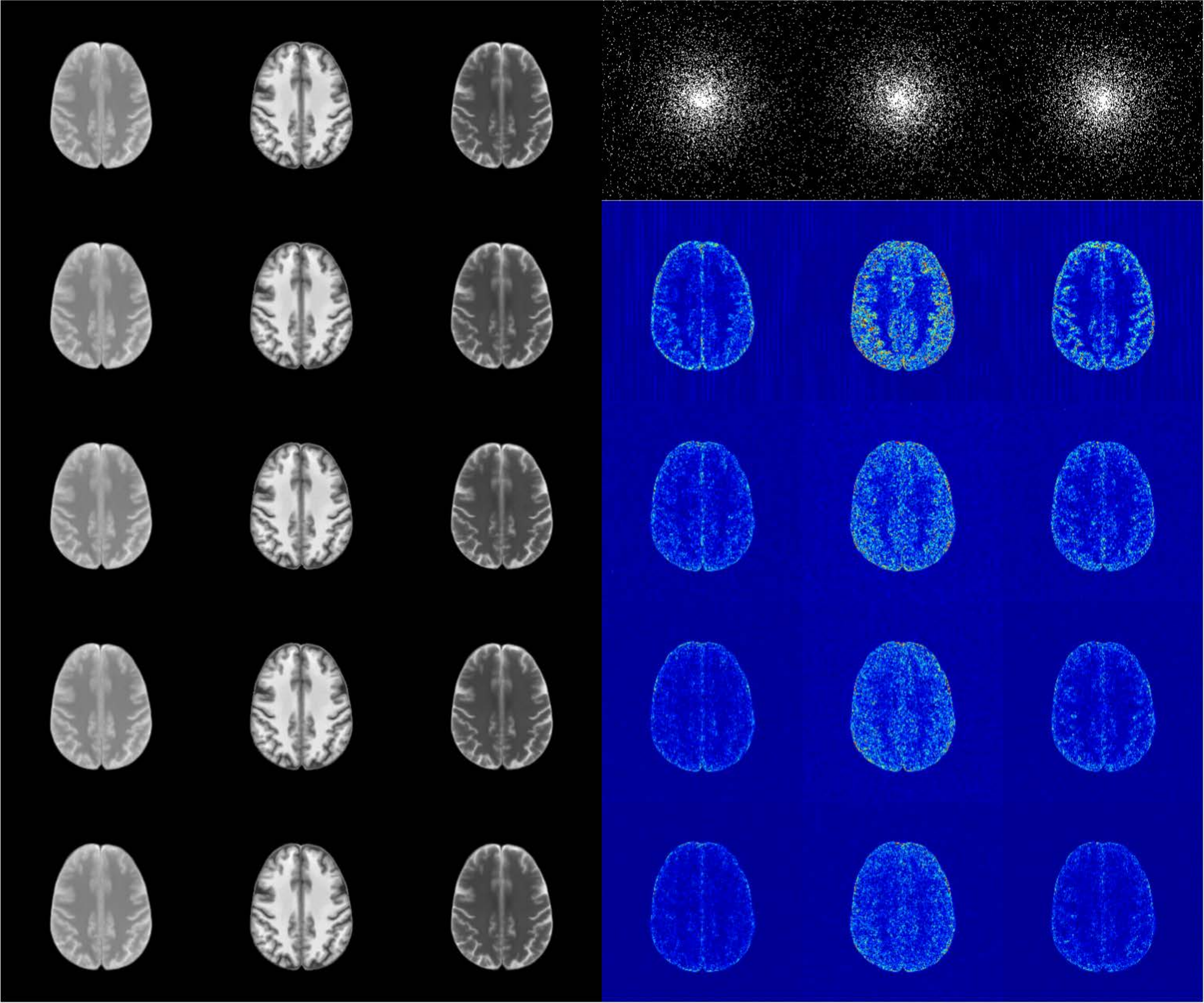}
   \caption{The reconstruction results for the comparison: the first row is the full-sampled MRI images with different 2D random $10\%$ masks. From second row to fifth row, we show the FCSA-MT, DIRN-5B, DFSN-5B and DISN-5B on a testing multi-contrast MRI data in SRI24 datasets. From left to right is the PD, T1 and T2 reconstructions respectively. The display of the error maps range from 0 to 0.1}
\label {figure9}
\end{figure}

On the SRI24 datasets, we compare the proposed DISN-5B model and its more basic versions DIRN-5B and DFSN-5B model with single-contrast MRI methods PANO \cite{11}, GBRWT \cite{13} and state-of-the-art multi-contrast methods, such as BCS \cite{23} and FCSA-MT \cite{27}, using three different $1$D Cartesian masks with the same sampling ratio of $20\%$. The parameter setting of the non-deep optimization models has been adjusted to optimal. The reconstructions and error residual images are shown in Figure \ref{figure7}. We see that the visual quality of DISN outperforms other methods, providing better preservation for structure details. This is supported by objective quantitative measures of PSNR and SSIM, which we show in Figures \ref{figure8a} and \ref{figure8b}. We find that a plain deep model like DIRN and DFSN already achieves good performance on the task, while the proposed DISN model achieves the best performance. We further test DISN using three different $2$D random masks with sampling ratio of $10\%$, and compare with FCSA-MT, DIRN and DFSN. These reconstruction results are shown in Figure {\ref{figure9}}. The experiment proves the DISN model can be well generalized to different sampling patterns with different under-sampling ratios.

\begin{center}
\begin{figure*}
\begin{center}
   {\label {figure10a} \includegraphics[width=0.76\textwidth]{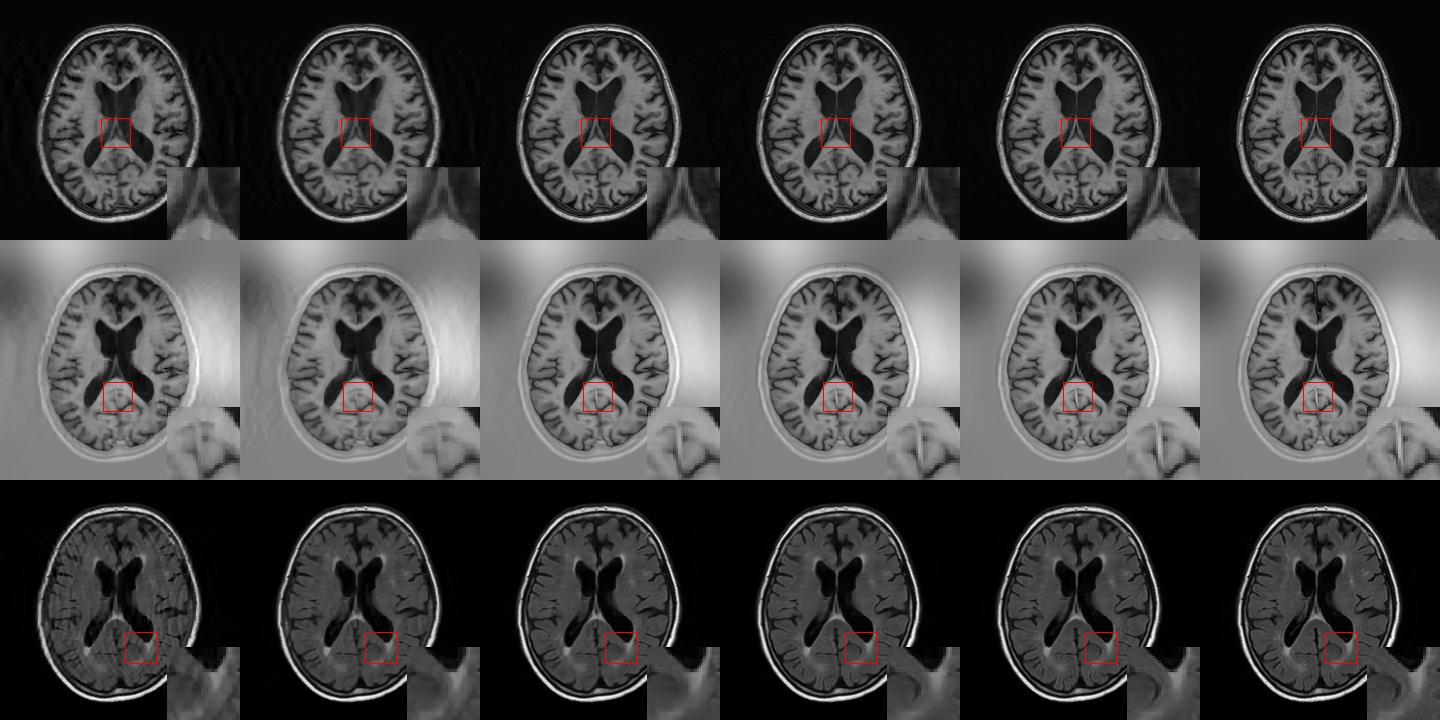}}
   {\label {figure10b} \includegraphics[width=0.76\textwidth]{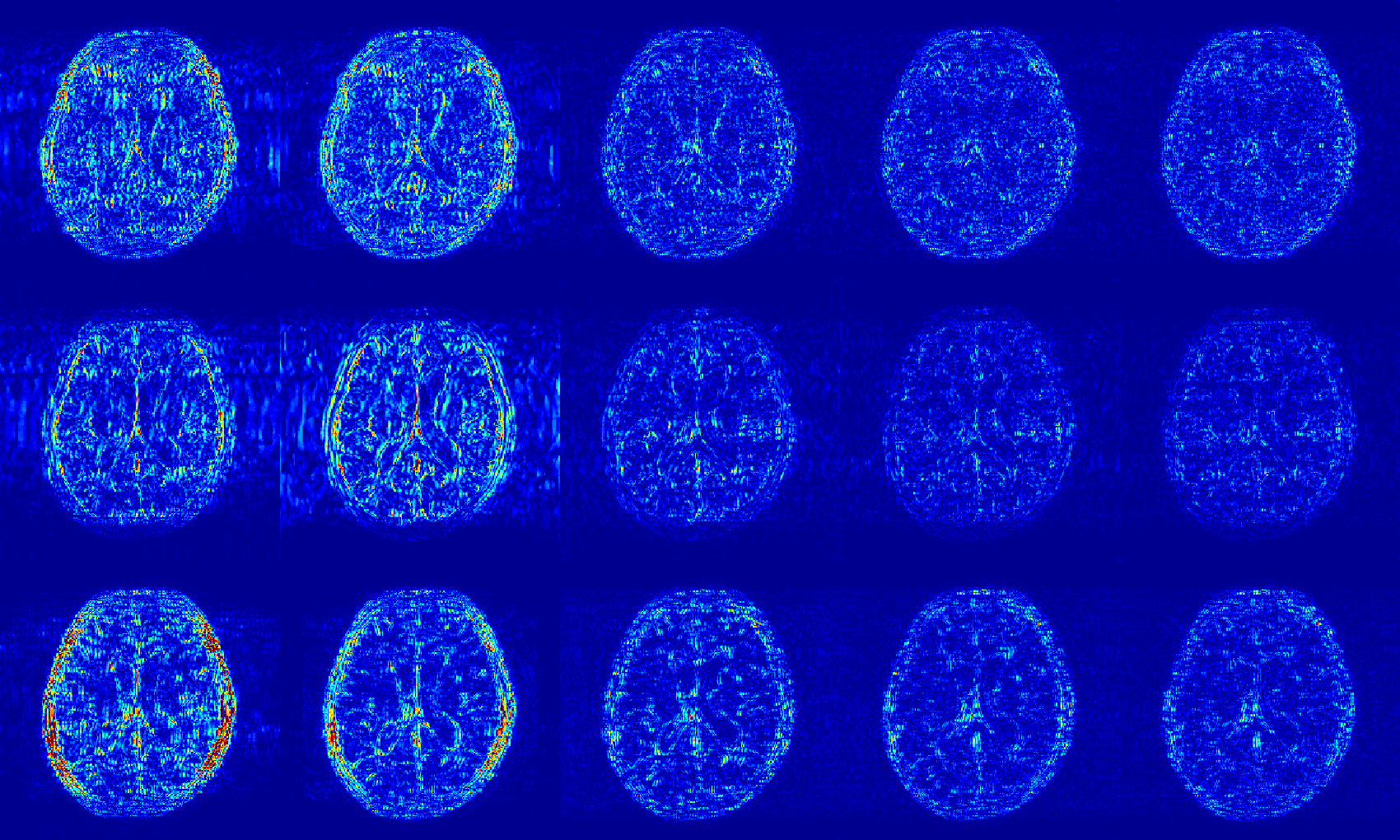}}
   \caption{In the first row (T1 contrast), second row (T1-IR contrast) and third row (T2-FLAIR contrast), we show the MR images of FCSA-MT, GBRWT, DIRN-5B, DFSN-5B, DISN-5B and full-sampled from left to right on a testing multi-contrast MRI data in MRBrainS benchmark. In the last three rows, we show the corresponding reconstruction error maps. The error display ranges from 0 to 0.15.}
\label {figure10}
\end{center}
\end{figure*}
\end{center}

\subsection{The model comparisons on the MRBrainS datasets}

\begin{table*}[t]
\centering
\caption{\scriptsize The averaged PSNR (dB) and SSIM of different CS-MRI methods on the test dataset. We also give the evaluation index for the regions of WML.}
\label{Table1}
\large
\begin{tabular}{|c|c|c|c|c|c|c|}
\hline
Regions   & \multicolumn{3}{c|}{Whole Brain}        & \multicolumn{3}{c|}{Regions of WML}     \\ \hline
Contrasts & T1          & T1-IR       & T2-FL    & T1          & T1-IR       & T2-FL    \\ \hline
BCS       & 28.77/0.854 & 30.92/0.922 & 29.40/0.788 & 27.80/0.816 & 28.27/0.852 & 29.39/0.851 \\ \hline
PANO      & 32.82/0.928 & 32.81/0.953 & 33.48/0.926 & 30.83/0.893 & 29.77/0.901 & 31.89/0.902 \\ \hline
GBRWT     & 33.16/0.939 & 33.10/0.958 & 33.91/0.943 & 31.00/0.896 & 30.17/0.909 & 32.07/0.901 \\ \hline
FCSA-MT   & 32.59/0.934 & 34.73/0.967 & 31.57/0.915 & 31.12/0.899 & 32.22/0.928 & 31.53/0.892 \\ \hline
DIRN-5B   & 36.48/0.967 & 37.54/0.978 & 34.52/0.935 & 33.87/0.935 & 33.70/0.944 & 32.90/0.916 \\ \hline
DFSN-5B   & 37.17/0.969 & 39.80/0.984 & 35.71/0.949 & 34.77/0.942 & 36.41/0.963 & 34.22/0.937 \\ \hline
DISN-5B   & \textbf{37.65/0.972} & \textbf{40.54/0.985} & \textbf{36.53/0.955} & \textbf{35.01/0.943} & \textbf{37.07/0.966} & \textbf{35.27/0.942} \\ \hline
\end{tabular}
\end{table*}

\begin{center}
\begin{figure*}
\begin{center}
\subfigure[PANO] {\label {figure11a} \includegraphics[width=0.2\textwidth]{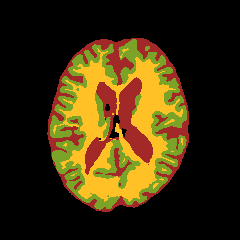}}
   \subfigure[GBRWT] {\label {figure11b} \includegraphics[width=0.2\textwidth]{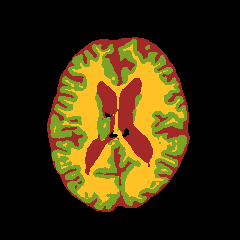}}
   \subfigure[FCSA-MT] {\label {figure11c} \includegraphics[width=0.2\textwidth]{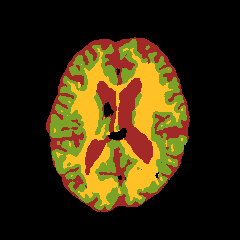}}
   \subfigure[DIRN-5B] {\label {figure11d} \includegraphics[width=0.2\textwidth]{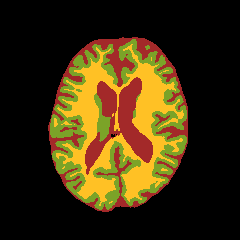}}
   \subfigure[DFSN-5B] {\label {figure11e} \includegraphics[width=0.2\textwidth]{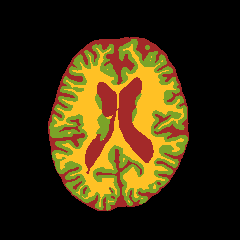}}
   \subfigure[DISN-5B] {\label {figure11f} \includegraphics[width=0.2\textwidth]{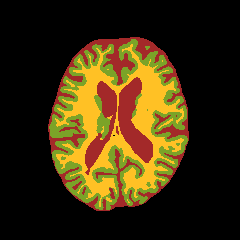}}
   \subfigure[Full-sampled] {\label {figure11g} \includegraphics[width=0.2\textwidth]{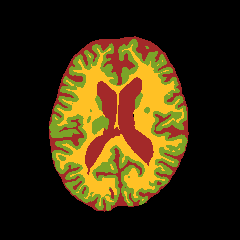}}
   \subfigure[Label] {\label {figure11h} \includegraphics[width=0.2\textwidth]{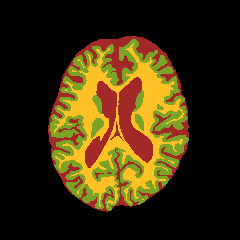}}
   \caption{The segmentation results produced by state-of-the-art U-Net architecture on compared single-contrast CS-MRI methods like PANO, GBRWT and multi-contrast CS-MRI methods like DIRN-$5$B, DFSN-$5$B and DISN-$5$B.}
\label {figure11}
\end{center}
\end{figure*}
\end{center}

\begin{table*}[]
\centering
\caption{\scriptsize The Dice Coefficients in $\%$ of the reconstructed T2-contrast MR images on the U-Net segmentation network. The segmented tissues include gray matter(GM), white matter (WM) and cerebrospinal fluid (CSF).}
\label{Table2}
\large
\begin{tabular}{|c|c|c|c|c|c|c|}
\hline
DC $\%$ & FCSA-MT & GBRWT & DIRN-5B & DFSN-5B & DISN-5B & Full-Sampled \\ \hline
GM      & 66.53   & 70.09 & 70.58   & 75.88   & \textbf{76.41}   & 77.93        \\ \hline
WM      & 77.66   & 79.30 & 79.90   & 84.56   & \textbf{85.03}   & 85.84        \\ \hline
CSF     & 73.85   & 75.81 & 77.96   & 80.53   & \textbf{80.97}   & 81.68        \\ \hline
\end{tabular}
\end{table*}

The standard scans in the SRI atlas contain no lesions of the brain and complex structural patterns, while the abnormalities show different appearances and diagnostic information across the different contrasts. Hence we also test our proposed model on the multi-contrast MRI datasets MRBrainS where the scans were acquired on patients with varying degree of white matter lesions (WML). We adopt 3 different 1D $20\%$ Cartesian masks for under-sampling as used in Figure \ref{figure7}.

In Table \ref{Table1} we observe the proposed DISN-5B still outperforms the compared state-of-the-art algorithms, followed by DFSN-5B. We show the reconstructions on a representative multi-contrast test MRI in Figure \ref{figure10}. From these experiments, we observe DISN model still works well in more complicated and diverse multi-contrast MRI settings. The employed deep neural network is flexible enough to model the structural similarities while distinguishing the structural differences across multi-contrast MRI, for example the white matter lesions regions are better recovered as shown in Table \ref{Table1} in the proposed DISN model, providing more reliable diagnostic information.

The MRI data are annotated with segmentation labels, and we test the reconstructed MR images produced by the compared CS-MRI models on the state-of-the-art well-trained medical image segmentation model called U-Net \cite{41} with pixel-wise cross-entropy as loss function. The model is trained with full-sampled MRI and label pairs. We use the widely-used Dice Coefficients (DC) as the objective index to evaluate the segmentation performance. The averaged objective DC results are shown in Table \ref{Table2} (DC index is in percent and higher score means better segmentation). Also, we show the subjective segmentation comparisons in Figure \ref{figure11}. We observe the better reconstruction produced by DISN model also leads to more accurate segmentation, which is near the upper bound of segmentation performance provided by the segmentation of full-sampled MR images on the U-Net model. The proposed multi-contrast MRI reconstruction model DISN can bring significant benefits in the medical image analysis.

\subsection{The model comparisons on the NeoBrainS datasets}

\begin{center}
\begin{figure}
\begin{center}
   {\label {figure12a} \includegraphics[width=0.5\textwidth]{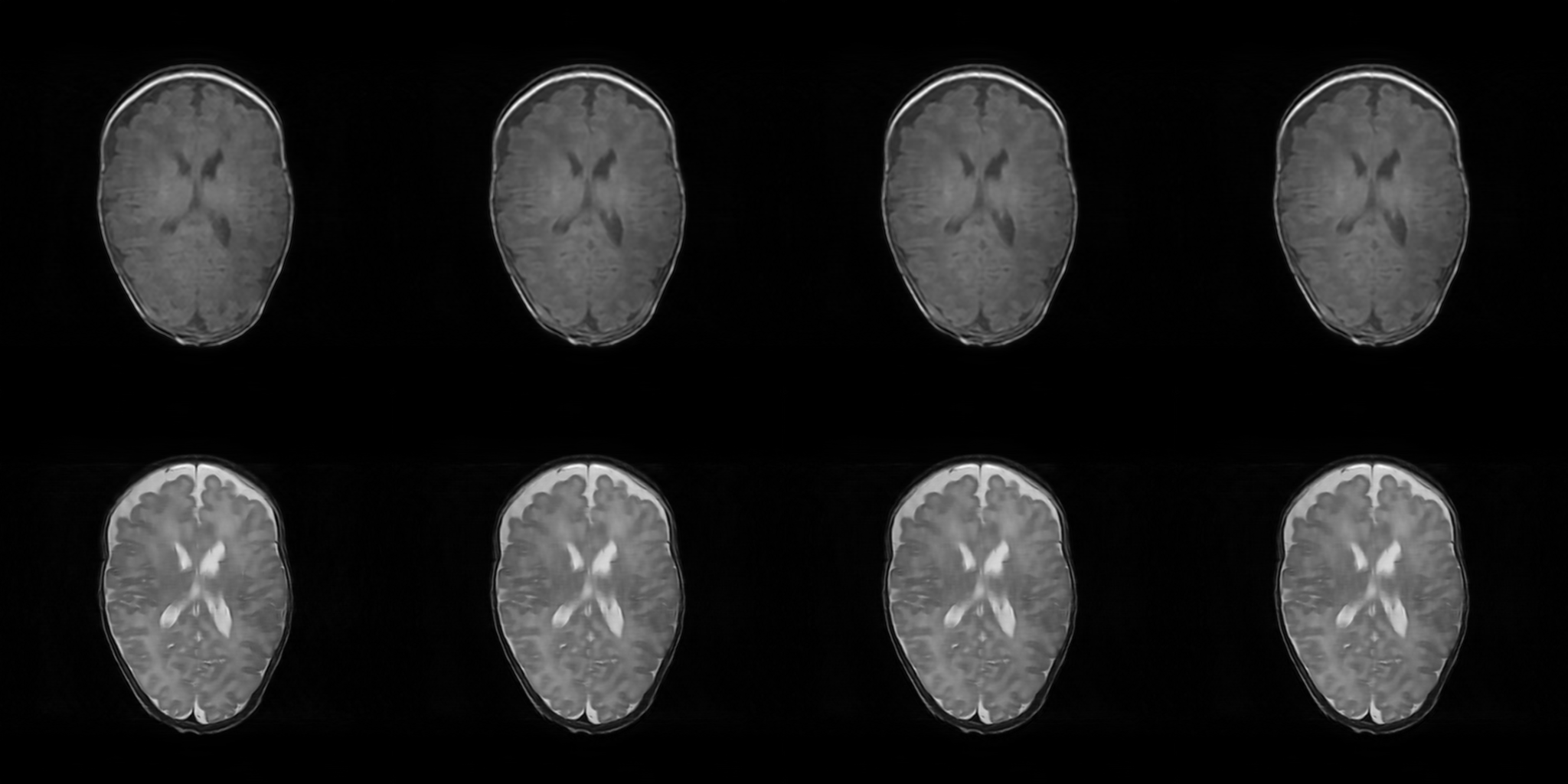}}
   {\label {figure12b} \includegraphics[width=0.5\textwidth]{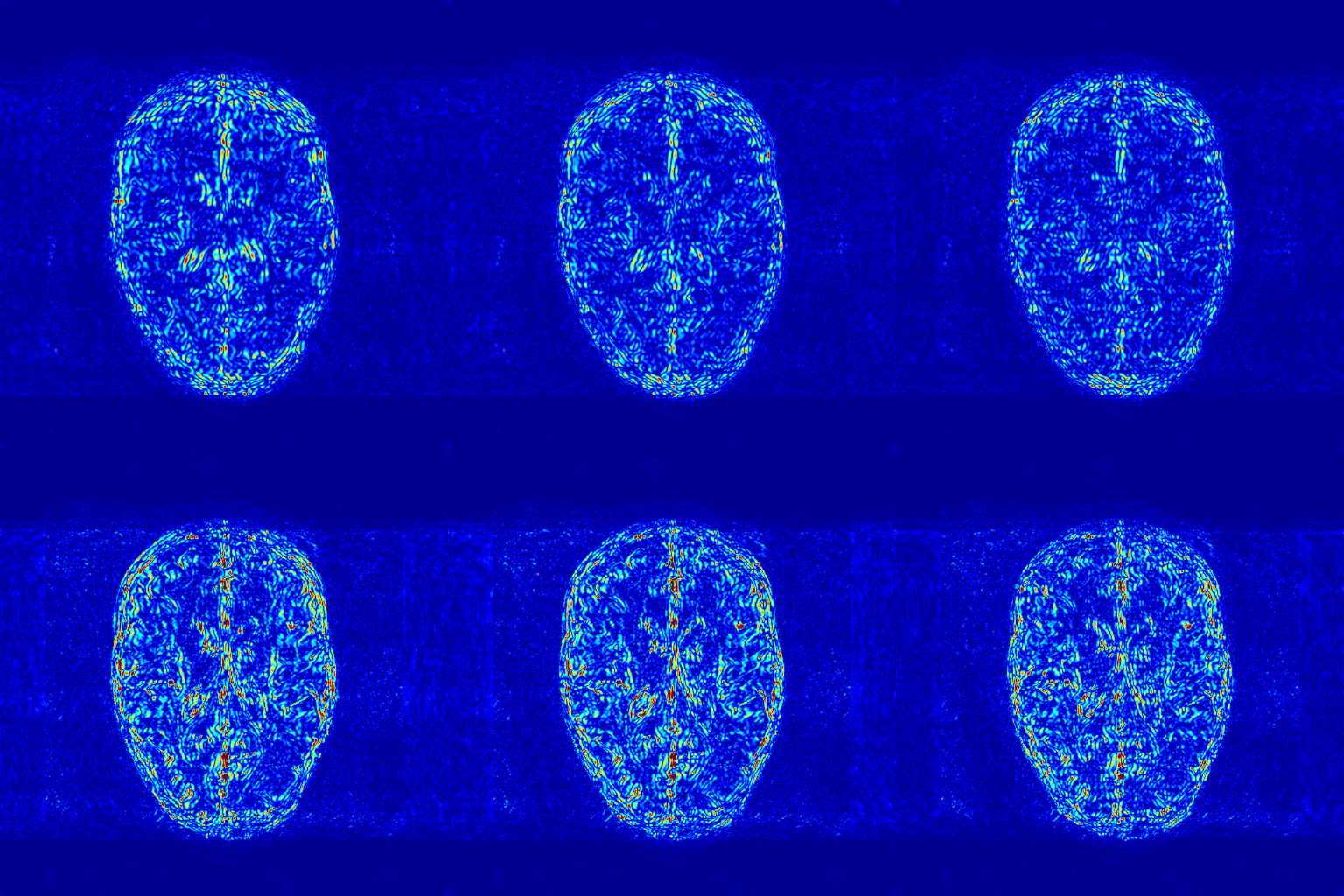}}
   \caption{In the first row (T1 contrast), second row (T2 contrast), we show the MR images of DIRN-5B, DFSN-5B, DISN-5B and full-sampled from left to right on a testing multi-contrast MRI data in NeoBrain benchmark. In the last three rows, we show the corresponding reconstruction error maps. The error display ranges from 0 to 0.1.}
\label {figure12}
\end{center}
\end{figure}
\end{center}

\begin{figure}
\begin{center}
   {\includegraphics[width=0.8\columnwidth]{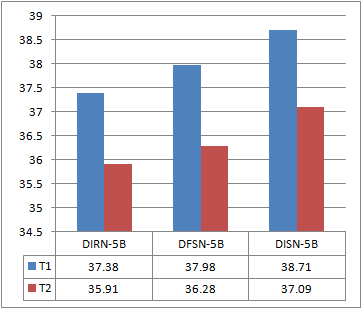}}
   {\includegraphics[width=0.8\columnwidth]{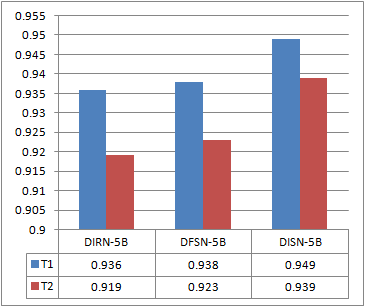}}
   \caption{The PSNR and SSIM of deep multi-contrast CS-MRI inversion algorithms averaged over the test images in the NeoBrainS benchmark.}
\label {figure13}
\end{center}
\end{figure}

Besides the multi-contrast MRI data MRBrainS benchmark acquired on patients (age $>$ 50), we also test the proposed multi-contrast MRI reconstruction model on the neonatal brain MRI in the NeoBrainS benchmark. The neonatal brains grow rapid and develop a wide range of cognitive and motor functions, which is critical in many neurodevelopmental and neuropsychiatric disorders, such as schizophrenia and autism. The DIRN-$5$B, DFSN-$5$B and DISN-$5$B are trained and tested on the training datasets in NeoBrainS benchmark with $10\%$ Cartesian under-sampling mask of the size $512\times512$.  We show the reconstructed MRI images of DIRN-$5$B, DFSN-$5$B and DISN-$B$B and their corresponding error maps in Figure \ref{figure12}. We observe the DISN-5B again achieves the optimal reconstruction quality. In Figure \ref{figure13}, we give the averaged PSNR and SSIM evaluation of the three compared deep models, which is consistent with the visual assessments.

The experimental results on the three different multi-contrast MRI datasets prove the proposed DISN model can also be well generalized to standard brain MRI datasets, brain MRI datasets with pathological abnormalities and neonatal brain MR datasets with different under-sampling patterns and under-sampling ratios.

\section{Discussion}

\subsection{Converge Analysis}

In Figure \ref{figure14} below, we show the training loss curve on the SRI24 atlas datasets as a function of iteration with masks from Figure \ref{figure7}. We observe the convergence for these deep models is relatively fast, and DISN gives a network with better training loss.

\begin{figure}
\begin{center}
{\includegraphics[width=1\columnwidth]{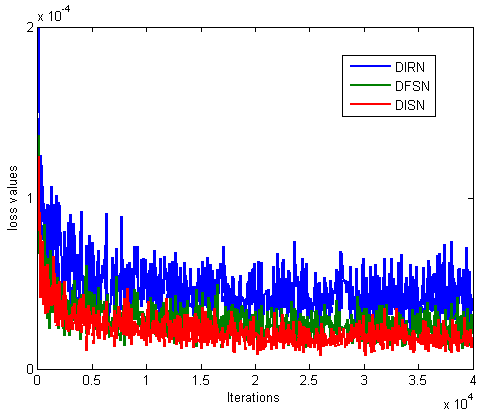}}
   \caption{The curves for the loss function value of DIRN-5B, DFSN-5B and DISN-5B.}
\label {figure14}
\end{center}
\end{figure}

\subsection{Network size}

\begin{figure}
\begin{center}
   { \includegraphics[width=1\columnwidth]{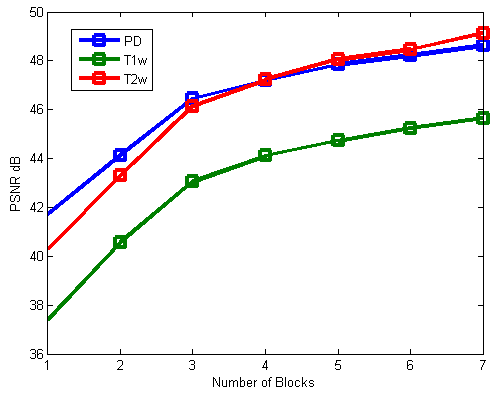}}
   \caption{PSNR curves as a function of the number of blocks.}
\label {figure15}
\end{center}
\end{figure}

Next we discuss DISN model performance by adjusting the number of cascaded blocks from $1$ to $7$ and give these results in Figure \ref{figure15} on the SRI24 datasets. We find as the number of blocks increases, the network performance steadily increase with smaller marginal improvement, while the DISN-$5$B model already achieves the state-of-the-art performance in multi-contrast CS-MRI reconstruction.

\subsection{Testing running time}

In Table \ref{Table3} we compare the running time for different models at testing time on the SRI24 datasets. For the optimization-based single- and multi-contrast MRI methods, additional optimizations are required on these test images, making processing of a new MRI more time-consuming. On the other hand, for DIRN-5B, DFSN-5B and DISN-5B, the reconstruction is much faster because the model is feed-forward and no iterations are required.

\begin{table}[h]
\caption{Test time comparison (deep models are for 5 blocks).\vspace{2pt}}
\label{Table3}
\scriptsize
\begin{tabular} {|l|c|c|c|c|c|l|}
\hline
BCS & PANO & GBRWT & FCSA-MT & DIRN & DFSN & DISN \\ \hline
30min & 19.6s & 64.6s & 5.7s & 0.17s & 0.11s & 0.18s \\ \hline
\end{tabular}
\end{table}

\subsection{Non-registration environment}

The multi-contrast MRI datasets used in this work have already undergone registration, i.e., been made to overlap as well as possible. However, in real MRI scenarios such accurate registration is not always realistic. For traditional optimization-based multi-contrast MRI methods such as FCSA-MT, this registration must be strictly satisfied because of the rigid sparsity assumption in these models. However, for the proposed DISN the trained network is quite robust to the shifts that are normal in the real-world MRI scanning process.

\begin{figure}
\begin{center}
   {\includegraphics[width=1\columnwidth]{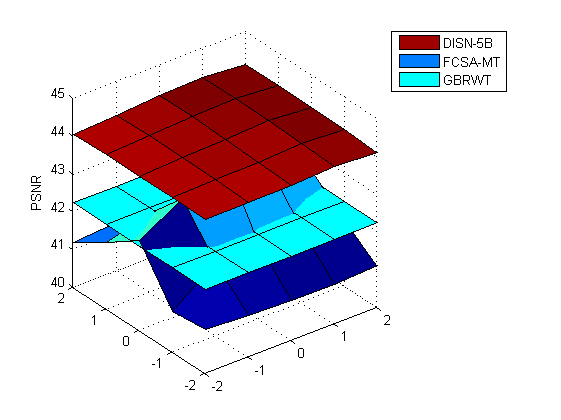}}
   \caption{DISN is robust to non-registration environments compared with conventional single-contrast and multi-contrast MRI in PSNR. The bottom plane indicates the degree of the pixel shift each direction while the vertical axis indicates the model performance in PSNR.}
\label {figure16}
\end{center}
\end{figure}

In this experiment, we take the SRI24 datasets for example and train the DISN-5B model with randomly shifted MRI data pairs in the small range within 2 pixels in all directions.
%of 1 pixel to 5 pixels.
We then test the DISN model on the position-fixed PD, T1 and position-shifted T2 data in the test datasets. The T2 data is also shifted within the 2 pixels in all directions.
%from 1 pixel to 5 pixel.
(We use the under-sampling masks shown in Figure \ref{figure7}.) Since FCSA-MT is done \textit{in situ}, there is no retraining required using shifted examples as necessary with DISN. However, in comparison between re-trained DISN and FCSA-MT, we observed that DISN is more robust to these pixel shifts. This is shown for the T2 reconstruction as a function of pixel shift in Figure \ref{figure16}.

We observe the FCSA-MT with well-registered MRI pairs outperforms the GBRWT, which is the state-of-the-art single-contrast CSMRI method only performing on the shifted T2 data, while the performance of FCSA-MT decreases dramatically when the shifting becomes severe. The DISN model steadily outperforms FCSA-MT and GBRWT regardless of the shifting. The simulation experiments show the proposed DISN model has the application potential in clinical MRI scenarios.

\section{Conclusion}
We have proposed the first deep models for the multi-contrast CS-MRI inversion problem. The model consists of densely cascaded inference blocks each containing a feature sharing unit and data fidelity unit. The feature sharing strategy can significantly reduce the number of parameters while still obtaining excellent model performance by virtue of the structural similarity across the multiple contrasts. The dense connection helps to share information across the blocks in a computationally efficient way. The experiments on different multi-contrast MRI datasets demonstrate that DISN achieves state-of-the-art performance in imaging quality and speed, bringing benefits to the later medical image analysis stage. Furthermore, its robustness to the non-registration environment shows potential for real multi-contrast MRI application.

\section*{Acknowledgment}

This work was supported in part by the National Natural Science Foundation of China under Grants 61571382, 81671766, 61571005, 81671674, U1605252, 61671309 in part by the Guangdong Natural Science Foundation under Grant 2015A030313007, in part by the Fundamental Research Funds for the Central Universities under Grant 20720160075, 20720180059, in part by the National Natural Science Foundation of Fujian Province, China under Grant 2017J01126.

\bibliographystyle{IEEEtrans}
\bibliography{egbib}

\end{document}